\documentclass[10pt, conference, compsocconf]{IEEEtran}

\usepackage{ifthen}
\usepackage[utf8]{inputenc}
\usepackage{lmodern} 
\usepackage[T1]{fontenc}
\usepackage{fancyhdr}

\usepackage{cite}
\usepackage{hyperref}

\usepackage[pdftex]{graphicx}
\usepackage{float}
\usepackage[table,xcdraw]{xcolor}

\usepackage[cmex10]{amsmath}

\usepackage{array}
\usepackage{mdwmath}
\usepackage{mathtools}
\usepackage{mdwtab}
\usepackage{eqparbox}
\usepackage{booktabs}
\usepackage{multirow}
\usepackage{multicol}

\usepackage{fixltx2e}
\usepackage{stfloats}

\begin{document}

\title{Improving tracking with a tracklet associator}

\author{\IEEEauthorblockN{Rémi Nahon, Guillaume-Alexandre Bilodeau and Gilles Pesant}
\IEEEauthorblockA{Department of Computer Engineering and Software Engineering\\
Polytechnique Montréal, Montréal, Canada \\
nahon.remi@polymtl.ca, gabilodeau@polymtl.ca, gilles.pesant@polymtl.ca}
}

\maketitle

\begin{abstract}

Multiple object tracking (\emph{MOT}) is a task in computer vision that aims to detect the position of various objects in videos and to associate them to a unique identity. 
We propose an approach based on Constraint Programming (\emph{CP}) whose goal is to be grafted to any existing tracker in order to improve its object association results. 
We developed a modular algorithm divided into three independent phases. The first phase consists in recovering the tracklets provided by a base tracker and to cut them at the places where uncertain associations are spotted, for example, when tracklets overlap, which may cause identity switches. In the second phase, we associate the previously constructed tracklets using a Belief Propagation Constraint Programming algorithm, where we propose various constraints that assign scores to each of the tracklets based on multiple characteristics, such as their dynamics or the distance between them in time and space. Finally, the third phase is a rudimentary interpolation model to fill in the  remaining holes in the trajectories we built.
Experiments show that our model leads to improvements in the results for all three of the state-of-the-art trackers on which we tested it (3 to 4 points gained on \emph{HOTA} and \emph{IDF1}). 
\end{abstract}


\ifCLASSOPTIONpeerreview
\begin{center} \bfseries EDICS Category: 3-BBND \end{center}
\fi
%
\IEEEpeerreviewmaketitle

\section{Introduction}
Multiple object tracking (\emph{MOT}) aims to detect and affect a unique identity to various objects in video sequences. 
It is in many cases solved by the \emph{tracking- by-detection} paradigm, which consists in separating the problem into two distinct tasks, detection and association (which we do in this work), but can also be solved by  \emph{tracking-by-regression}, a method that performs these two actions in parallel for short-term associations. While recent advances in machine learning (\emph{ML}) have led to a huge performance gain for the detection phase in MOT, the association phase remains a challenge, especially because of its combinatorial complexity. 

In this paper, to better deal with the combinatorial complexity in the association phase, we propose a module based on Constraint Programming (\emph{CP}) whose goal is to be grafted to any existing tracker in order to improve its object association results. It can be applied to methods that are from either the \emph{tracking-by-detection} or \emph{tracking-by-regression} framework since our method addresses long-term data association at the tracklet level. Our proposed model is divided into three independent phases. The first phase, called \emph{TrackletCutter} consists in recovering the tracklets provided by a base tracker and to cut them in places where uncertain associations are spotted. In the second phase, called \emph{CP Associator}, we associate the previously constructed tracklets using a Belief Propagation Constraint Programming model, where we propose various novel constraints that assign scores to each of the tracklets based on multiple characteristics, such as their dynamics or the distance between them in time and space. Finally, the last phase is a rudimentary interpolation model to fill in the remaining holes in the trajectories we built.

In the experiments, we show the benefit of our method with improvements in the results for all three of the state-of-the-art trackers on which we have tested it (3 to 4 points gained on the \emph{HOTA} and \emph{IDF1} metrics).

\section{Background and related work}
MOT is a fast-growing field with many existing approaches whose performance has been greatly improved by the recent advances in machine learning. It can mostly be divided into two aspects (that are most often phases of resolution) : the object detection and the object association.

\subsection{Object Detection}

ML detection techniques have recently allowed great advances in the field of object detection.
This is the case, for example, of the detectors of the \emph{R-CNN} family \cite{RCNN} whose principle consists in extracting regions of interest (\emph{ROI}) and then, via convolutional neural networks (\emph{CNNs}), in inferring the main features of these ROI in order to find the class of an object and its exact position in the image. Improvements have been made to this method, in particular with the Faster \emph{R-CNN} detector (\emph{FRCNN}) \cite{FRCNN} for  which the region proposal method is itself a neural network called \emph{Region Proposal Network}. 
The typical output of a detector is usually provided in a $(x,y,w,h)$ format (where $(x,y)$ represent the spatial coordinates of the upper-left corner of the bounding box and $(w,h)$ its width and height). Such detections are typically the inputs of trackers. Since detection is not the subject of this article, we can just remember here that better detections lead to better tracking.  


\subsection{Object Association}

Given a set of detections at each frame, a \emph{MOT} method typically builds tracks by associating detections across frames. The tracker aims at affecting a single identifier to each object of interest. To efficiently perform the association phase, two questions arise: 
\begin{enumerate} 
    \item How to represent the detections in such a way as to be able to recognize and differentiate the different objects of interest? 
    \item How to efficiently explore the set of solutions and reach a satisfactory solution in a reasonable time?
\end{enumerate}

\begin{figure}[!t]
      \centering
        \includegraphics[width=.4\textwidth]{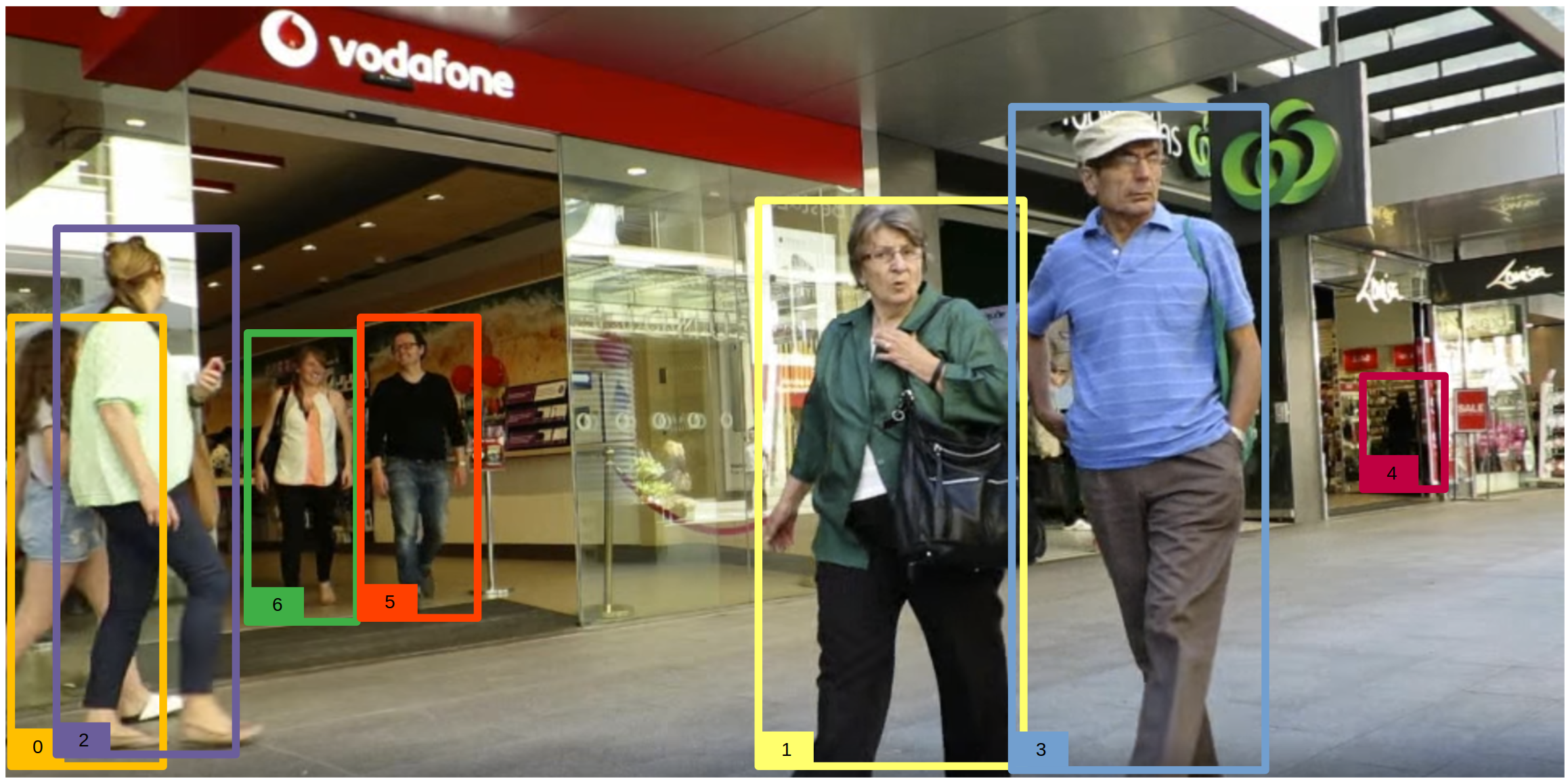}
        \caption{Typical tracking result for a frame in the MO17-09 sequence : each of the pedestrians is enclosed by a bounding box whose color corresponds to an identifier that should be kept by the same pedestrian during the entire sequence.}
        \label{fig:tra}
\end{figure}

\subsubsection{Models to describe the detections}
Regarding the representation of detections, for an efficient association, we generally seek to maximize one (or more) similarity metric between associated detections, and these metrics are computed from descriptors that are mainly divided into motion and appearance models.

\emph{Motion models} are generally based on the fact that the objects of interest being tracked meet a certain number of physical constraints concerning their speed of movement and deformation in the image. 
Because videos are often captured at more than 30 frames per second, the displacement in pixels of the classes of objects of interest is normally small between frames. We can consider what we will call \emph{positional models} which make the hypothesis that at the scale of successive frames, the object of interest followed is immobile, and of fixed size and shape. 
This allows the design of very simple and therefore very fast models such as the one on which much of our study is based, \emph{IOU-Tracker} \cite{IOU2017}, which consists in looking for the best associations of detections frame by frame by maximizing their \emph{IOU} (quotient of their intersection and union surfaces). This metric measures the superposition of two detections, and penalizes the differences in size and position. 
To increase the accuracy of this kind of \emph{positional model}, one can consider that the derivatives of these position and size values are fixed and use these to predict the position of the object in subsequent frames. This is the principle of the \emph{Kalman filter} \cite{Kalman} used for example in \emph{SORT} \cite{SORT}, one of the trackers on which we test our model. The tracking task can be simplified by keeping a model of the camera movement in parallel with the movement models of the objects of interest \cite{CMC}.

\emph{Appearance models} consist in the description of some visual characteristics of the objects of interest (or rather of their bounding boxes) and are based on the fact that an object of interest generally keeps a similar appearance through time (or changes only slightly and progressively anyway), which will allow to associate the detections with similar descriptors. These descriptors are generally based on the distribution of colors \cite{HOC}, gradients of intensities \cite{HOG} or more advanced methods such as covariance matrices and multiple kind of filters \cite{KCF}.
\emph{CNNs} can also be used to model appearance \cite{DAN}, as well as transformers \cite{MOTer_TransCtr} and Mixture Density Neural Networks \cite{trajeocc}. The method proposed by \emph{Tracktor} \cite{Tracktor}, on which we test our model, consists precisely in a regression of bounding boxes from one frame to another to extend the trajectories. \emph{CenterTrack} \cite{CenterTrack} works in a similar way but by working on the center of the detections, which allows it to be more robust to occlusions. For all these methods the regression is used for short-term association.

\subsubsection{Resolving the association problem}
Once the representation models have been chosen, we try to associate detections with one another to build our trajectories. The goal is then to assign a trajectory identifier to each detection to ensure that we can track every object of interest from its entry into the camera field of view (\emph{FOV}) to its exit of it.
Trackers can be divided into two categories regarding the way they process data. 

\emph{Online trackers} \cite{IOU2017, SORT} aim to build their trajectories in real-time and therefore work frame by frame in chronological order. The method they use must thus be incremental and build up tracks by adding detections to existing tracks at each new frame. The tracker must have a criterion to open a track (i.e. state that a new object entered the FOV), to close one (i.e. state that it has left the FOV) and to add a detection to a track. These kinds of trackers can only be based on information from past trajectories and current detections (that provide only static information) and cannot return to modify previously constructed trajectories.
    
\emph{Offline trackers} remove the real-time constraint and can thus be applied to the whole set of detections at once, and our work fits into this paradigm as it allows to consider the association problem as a global optimization problem. Even if it allows access to many new characteristics of the objects of interest and modes of resolution, this shift to \emph{offline} greatly increases the complexity of the problem. Different avenues are taken to resolve that difficulty. Some methods resolve it by still working frame by frame, whether it is with dynamic programming \cite{GOG} or by going through time in both directions \cite{TMOH} to make their results more robust than they would be online. Some choose to go from a local to a global scale, as in the \emph{H2T} tracker \cite{H2T}, which divides frames into small subsets, minimizes a sum of affinity functions to associate detections within them, and then resolves recursively the same problem on bigger and bigger associated sets until whole tracks are finalized. 

\section{Approach and methodology}
\subsection{Motivation} 

We build on the work of \cite{pineault_article} who was the first to apply \emph{CP} to multi-object tracking. The benefit of choosing \emph{CP} is that it is a very efficient method of formalizing and solving optimization problems \cite{pesant_cp}. This is the reason why we have decided to pursue in that direction. The choice of working on the association of pairs of detections was the main limitation of this previous work. Indeed this made the search space extremely vast and therefore the computation time quite high. This forced the decision to work in batches of frames and to strongly restrict the spatial distances between bounding boxes considered, which increases the sensitivity of the model to occlusions. 

To remedy this difficulty, our main idea is to work at the level of tracklets (sequences of detections) instead of individual detections. This greatly reduces the size of the problem. Indeed there are simple trackers that are extremely efficient in terms of execution time (notably \emph{IOU-Tracker} \cite{IOU2017} that manages to process up to 100,000 frames per second, making it hundreds of times faster than most of the other state-of-the-art trackers) and they are good at making associations in simple cases. We decided to start from them to capitalize on their speed and try to correct their errors a posteriori, especially those which intervene in the case of occlusions. This choice also allows us to increase the refinement of the association model, which can now be based on the characteristics of the tracklets, which are dynamic, and not only on those of the detections, which are static.  

Our model is also applicable to all kinds of trackers since we work from their outputs. Here we will speak of \emph{tracklet} for any association of detections provided by an initial tracker, whether we have cut it or not, and of \emph{trajectory} to speak either of the associations of tracklets that we constructed, or of the associations of detections provided by the ground truth representing a single object.

\subsection{A three-phase model}
Our model is divided into three modules : 
\begin{itemize}
    \item the \textbf{\emph{TrackletCutter}}  cuts the tracklets provided by the initial tracker where they intersect each other with a high degree of overlap;
    \item the \textbf{\emph {CP Associator}} is the original association model we propose here, based on a Belief Propagation Constraint Programming algorithm~\cite{miniCPBP};
    \item an \textbf{\emph{interpolation model}} fills the gap in trajectories by linearly interpolating these detections based on the ones at the edges of the gaps.

\end{itemize}
While the \emph{CP Associator} cannot be disabled (as it would prevent association), the other two modules are optional.

\subsection{TrackletCutter - Cutting tracklets on overlapping sections}
\begin{figure*}[!t]
        \centering
        \includegraphics[width=.95\textwidth]{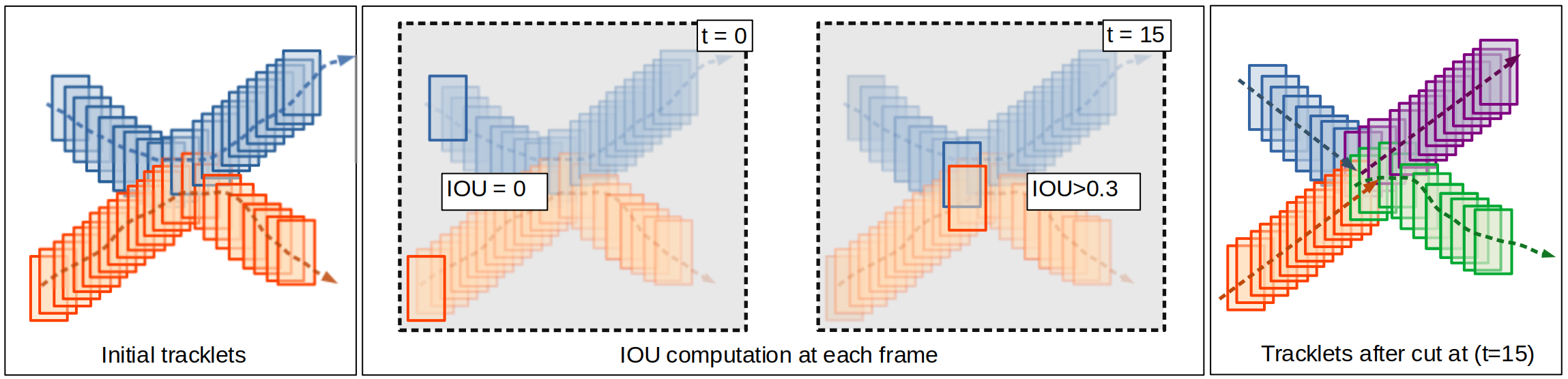}
        \caption{Workings of the $TrackletCutter$ with $T_{TC}$ = 0.3: Exploration of each frame one by one; as soon as the overlap of two bounding boxes reaches $T_{TC}$ their tracklets are cut.}
        \label{fig:tracklet_cutter}
\end{figure*}

While the sensitivity to occlusions of most trackers often leads to fragmentation (i.e. tracks being cut into multiple tracklets), we developed a module designed to separate tracklets that are at risk of containing multiple different objects of interest. As we know that this risk is at its highest when multiple objects of interest cross paths, it was decided to proceed as follows : as shown in Figure~\ref{fig:tracklet_cutter}, whenever in one frame two detections have an overlap (\emph{IOU}) that reaches a fixed threshold $T_{TC}$, the tracklets to which they belong are cut at that frame.

\subsection{Tracklet modeling}
\begin{figure}[!t]
    \centering
    \includegraphics[width=0.45\textwidth]{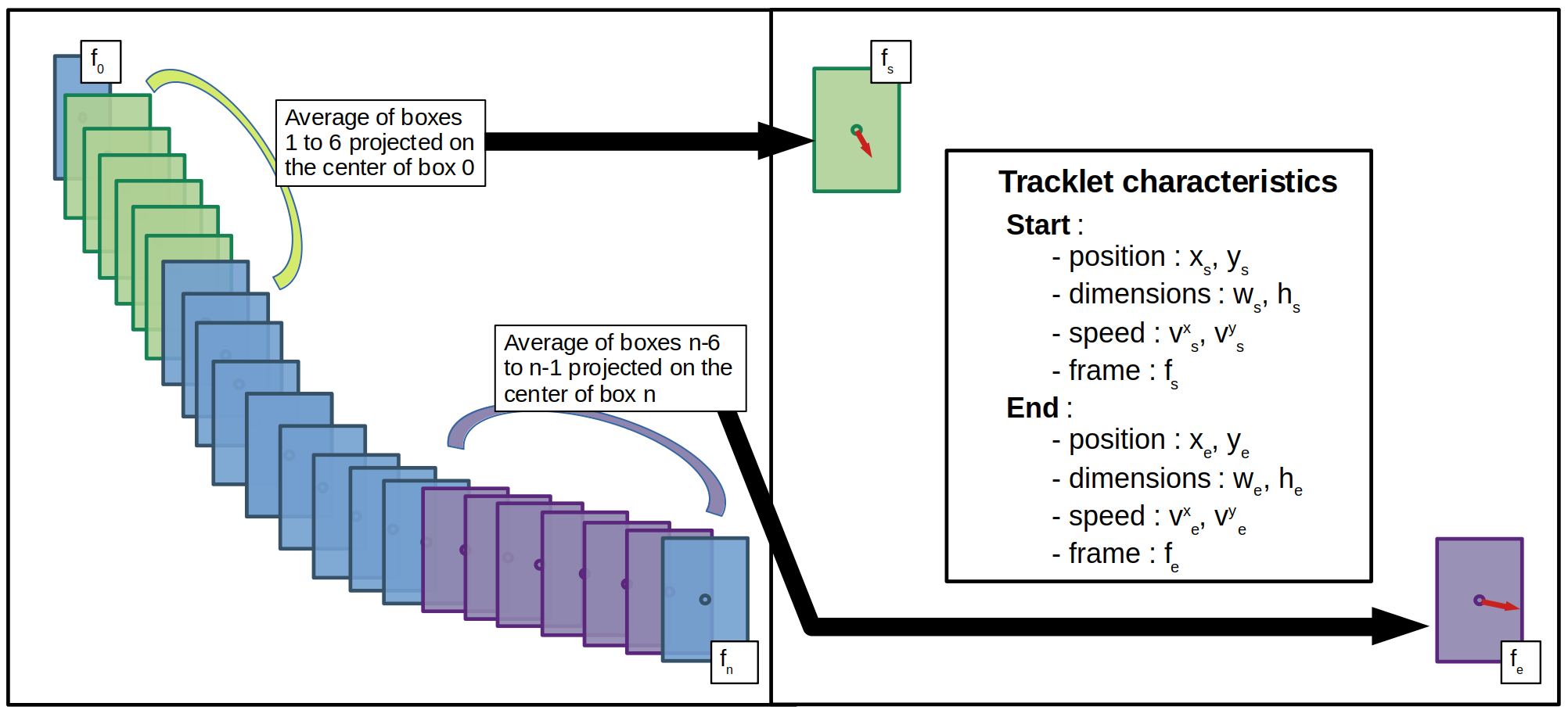}
    \caption{Construction of the two representative bounding boxes of a tracklet by averaging bounding boxes at its start and end.}
    \label{fig:avg}
\end{figure}

We define our set of tracklets as $T$ and each tracklet $t \in T$ is, as shown in Figure~\ref{fig:avg}, described by:
\begin{itemize} 
    \item a frame: $f_S$ (resp. $f_E$) the frame in which the first (resp. last) detection of the tracklet appears;
    \item a bounding box: ($x_S$,$y_S$,$w_S$,$h_S$) (resp. ($x_E$,$y_E$,$w_E$,$h_E$)) the mean of the spatial coordinates of the six bounding boxes following the first one (resp. preceding the last one) of the tracklet;
    \item a speed: ($v^x_S$,$v^y_S$) (resp. ($v^x_E$,$v^y_E$)) the mean speed of the centers of the the six bounding boxes following the first one (resp. preceding the last one) of the tracklet.
\end{itemize}

If $t$ is formed by fewer than ten detections, the averaging of the six first and last bounding boxes is not performed. The speed is then computed between the first (resp. last) two bounding boxes and the spatial coordinates are those of the first (resp. last) bounding box. However, doing this for each tracklet would have led us to put too much weight on the quality of these first and last bounding boxes that are by essence the least representative of the object of interest (as we can infer that the track has been separated at them because of some defect or occlusion), thus the choice of averaging the few following (resp. preceding) detections.  The choice of working on six bounding boxes is however arbitrary and could be refined in future works.

\subsection{Associating tracklets with Constraint Programming}
Once the tracklets have been modeled, the goal is to associate them using our \emph{CP Associator}. To do so, we need to model the problem following the CP paradigm, to define our use of constraints (both filtering and assigning marginals) and to define the methods we use to explore the solution space.
\subsubsection{Modeling the association problem in a \emph{CP} paradigm}
A CP model is given by a finite set of variables, each taking its value from a finite set called its domain.
Constraints are then specified on the combinations of values that these variables can take. This model defines a \emph{Constraint Satisfaction Problem} (\emph{CSP}). 
In our case, the tracklet association phase is modeled as:
\begin{itemize}
    \item $S$ our set of successor variables such that, for every tracklet $i$, $S(i)$ is its successor, meaning that $S(i)$ immediately follows $i$ in the same trajectory. 
    \item For every tracklet $i$, the domain $D(i)$ contains every tracklet that starts temporally after $i$ ends, and a stopping value (meaning that $i$ is the last tracklet of its trajectory).
    \item $C$ is our set of constraints, which we use to filter the domains of each successor and to affect a score to each tracklet-successor pair.
\end{itemize}
\subsubsection{Using constraints to filter}
Constraints are most often used to restrict the domains of variables. We use them in that fashion to ensure the following characteristics for our trajectories:

\paragraph{allDifferent}
No detection should be found in multiple trajectories and therefore no tracklet should be assigned to multiple trajectories. To accomplish that goal, we used the \emph{allDifferent} constraint. Applied to the whole set of variables, it ensures that no two variables are assigned the same value.
It is given by:

\begin{equation}
    \forall (i,j) \in T^{2} \, | \, i\neq j, \, S(i) \neq S(j)
\end{equation}

\paragraph{Temporal consistency}
As we suppose in this part of our model that our tracklets are perfect (namely that each detection in a specific tracklet belongs to a single object), there should be no overlap in time between tracklets affected to a single trajectory. Therefore, we define the temporal consistency constraint as follows: 

\begin{equation}
    \forall (i,j) \in T^{2},  j \in D(i) \Rightarrow f_{E}(i) < f_{S}(j)
\end{equation}

\subsubsection{Score-based constraints}
We also propose to use constraints in a different manner, that is to assign a score to each pair $(t,s)$ (where $t$ is a tracklet and $s \in D(t)$ is a successor) based on a given characteristic $c$. Each constraint is based on a distance $c(t,s)$ between $t$ and $s$. These different kinds of distances (that can be, as we will see below, temporal or spatial for example) are then transformed into scores $S_{c}(t,s)$ ranging from 0 to 1 where: 
\begin{itemize}
    \item $S_{c}(t,s) = 0$ leads to the immediate removal of $s$ from $D(t)$ the domain of the successors of $t$.
    \item $S_{c}(t,s) = 1$ leads to the immediate assignment of $s$ to $S(t)$ i.e. $s$ becomes the successor of $t$. 
    \item $S_{c}(t,s_1) \geq S_{c}(t,s_2)$ means that according to characteristic $c$, $s_1$ is a more likely candidate than $s_2$ to be the successor of $t$.
\end{itemize}

\begin{figure}[!t]
        \centering
        \includegraphics[width=.45\textwidth]{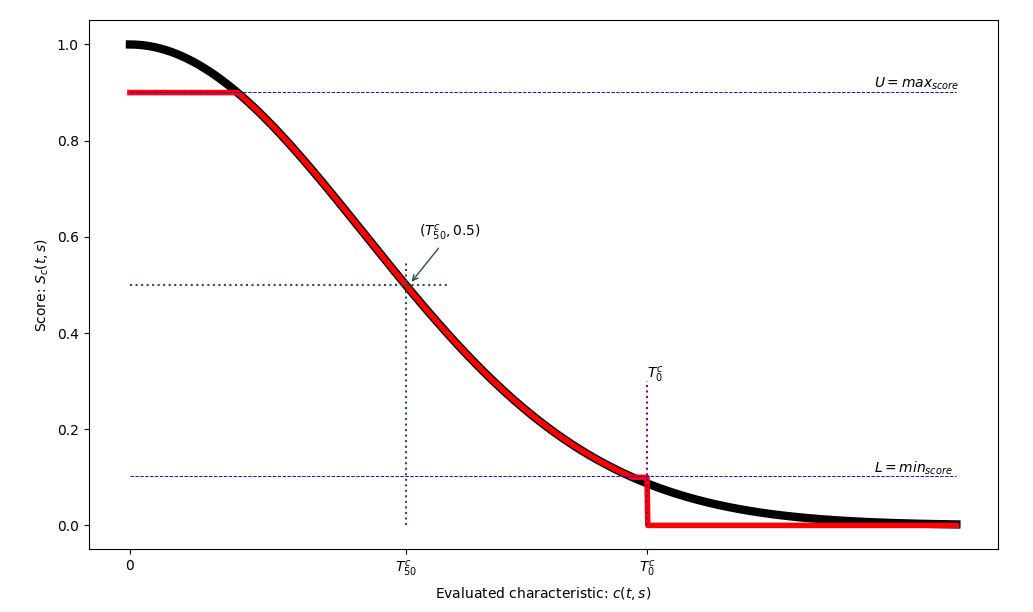}
        \caption{Representation of the computation of a constraints score : the thick black line represents $S$ (gaussian of peak 1, mean 0 and standard deviation $\frac{1}{2ln(2)}T^c_{50}$), the red line represents the real score: bounded above by U and below by L, falls to 0 when reaching $T^c_0$. To enhance visibility we set U = 0.9 and L = 0.1. Typically we set $L = 10^{-6}$ and $U = 1-L$.}
        \label{fig:score}
\end{figure}

\paragraph{Building scores}
Each score-based constraint is assigned three thresholds that will help define their behavior:
\begin{itemize}
    \item $T^c_{50}$: value of the distance for which we set the score at $50\%$ (i.e. $0.5$).
    \item $T^c_{end}$: value of the distance between the tracklet $t$ and the fictional tracklet which represents the stopping of the trajectory at $t$.  It is indeed essential to compare the set of possible successors to the possibility of associating with none of them (which would mean that $t$ is that last tracklet of its trajectory).
    \item $T^c_0$: value of the distance beyond which the examined successor $s$ is removed from $D(t)$. This threshold may or may not be activated, but if it is and it is reached for a given pair $(t,s)$, $S_c(t,s)$ is then equal to $0$ and $s$ is removed from $D(t)$.
\end{itemize}
As shown in Figure~\ref{fig:score}, as long as $c(t,s) \leq T^c_0$ (where it automatically falls down to 0), the score of the association of $t$ and $s$ according to the characteristic $c$, $S_c(t,s)$, is calculated as follows. We compute 
    \begin{equation}
    S = exp(\frac{-c(t,s)^2}{2\sigma_c ^2}) 
    \end{equation}
    where 
    \begin{equation} 
    \sigma_c = \frac{T^c_{50}}{2ln(2)}
    \end{equation}
    so that $S = 0.5$ for a distance of $T^c_{50}$.
     Finally the actual score is bounded as follows 
    \begin{equation}
        S_c(t,s) = max(L, min(S,U))
    \end{equation}
    where $L$ is the lower bound of the score and $U$ the upper bound (so that it does not reach 0 or 1). 

\paragraph{Constraint on time spacing}
The first kind of score-based constraint we developed favors a small temporal distance between a tracklet and its successor. For each pair $(t,s)$, we get the metric $td(t,s)$ ( where $td$ stands for \emph{time distance}) such that:
\begin{equation}
    td(t,s) = f_S(s) - f_E(t)
\end{equation}

\paragraph{Constraints on dynamics}
We also decided to built constraints that aim to maintain the trajectories as smooth as possible by minimizing the discontinuities in acceleration. For each pair $(t,s)$ we then obtain the metrics $ad(t,s)$ and $sd(t,s)$ (for \emph{angle difference} and \emph{speed norm difference}) such that: 
\begin{equation}
    ad(t,s)= \widehat{(\overrightarrow{v_E(t)},\overrightarrow{v_S(s)})}
\end{equation}
\begin{equation}
    sd(t,s)=\left | \overrightarrow{v_S(s)} \right | - \left | \overrightarrow{v_E(t)} \right |
\end{equation}

\begin{figure}[!t]
        \centering
        \includegraphics[width=.45\textwidth]{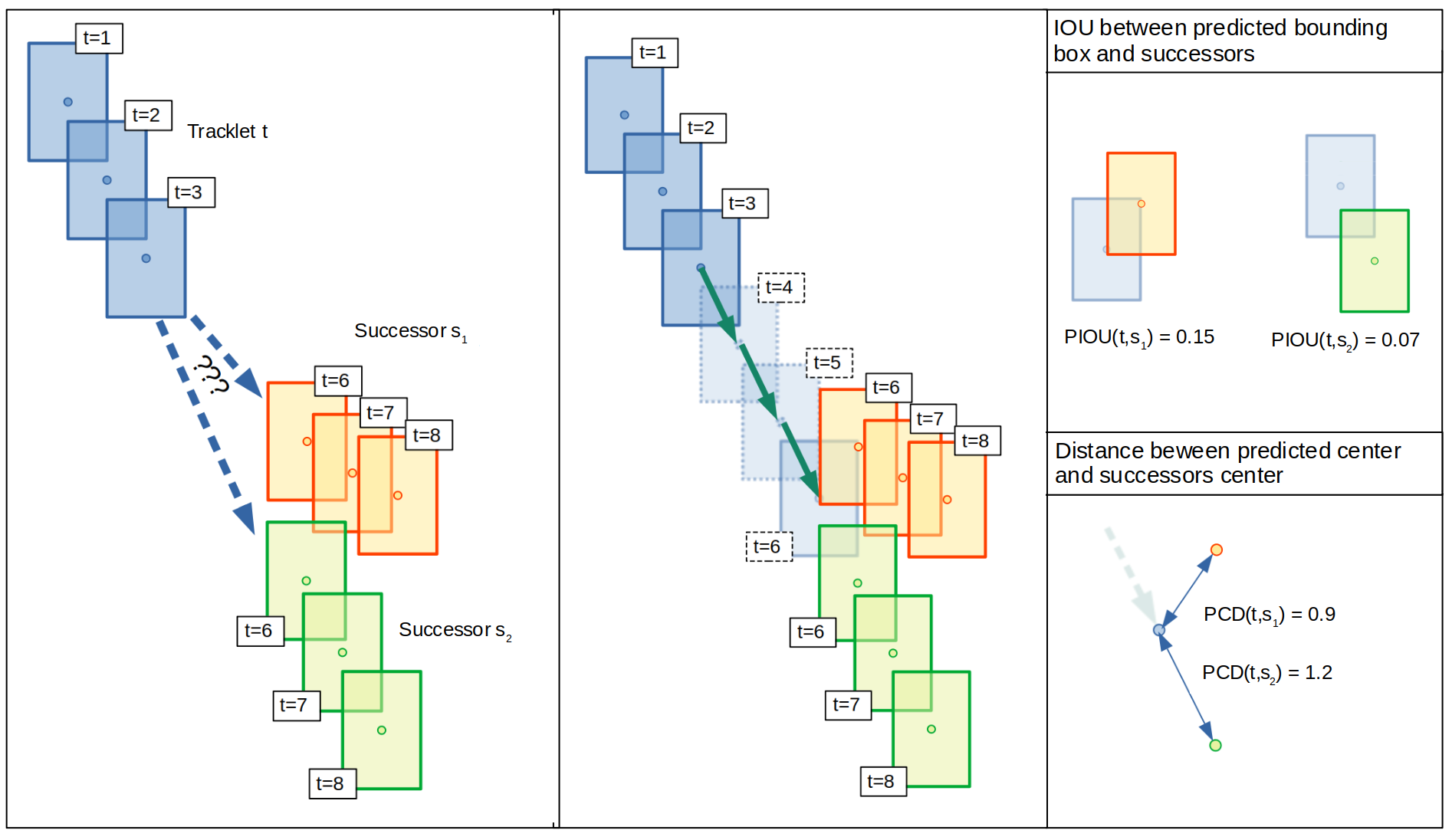}
        \caption{Computation of the characteristics evaluated by the \emph{predicted IOU} ($PIOU$) and \emph{predicted center distance} ($PCD$) constraints : the ending bounding box of the tracklet is projected onto the first frame of the successors candidates. The starting bounding boxes of the successors are compared with the projection based on their overlap and the distance between their centers.}
        \label{fig:piou_pcd}
        
\end{figure}

\paragraph{Constraints on a predicted position}
We can suppose that with the help of the aforementioned \emph{time distance} constraint that the tracklets most likely to be associated are those that are not too temporally distant. As the shorter the time interval, the less the object of interest can change its speed, direction and even position in the image, we decided to add a constraint on a predicted position. Following the example of a \emph{Kalman filter} \cite{Kalman}, we consider that projecting a bounding box using its speed onto subsequent frames not too far apart is a relatively efficient prediction mechanism. Therefore we propose two constraints, which are described in Figure~\ref{fig:piou_pcd} : they compare the predicted position of the considered object and the evaluated successor by \emph{IOU} or by the distance between their centers (which is largely considered to be more robust to occlusions \cite{CenterTrack}).

\subsubsection{Adaptation to video sequences properties}
As the video sequences we work with have different characteristics regarding dimensions and capture speed (measured in frames per second or \emph{FPS}), we chose to adapt the parameters of the score-based constraints. Indeed, affecting a \emph{time distance} score of $0.5$ to a pair of tracklets separated by six frames has a very different meaning for a sequence of $12$ FPS compared to one four times faster. Therefore we decided to adapt
the \emph{time distance} constraint to the FPS of the sequence,
    the \emph{predicted center distance} constraint to the size of the image (represented by proxy by the length of the diagonal of the image),
and the \emph{speed norm difference} constraint to the FPS and the diagonal.

\subsubsection{Marginals}
Once each of the successor domains $D(t)$ have been restricted by the activated relevant constraints, the remaining $(t,s)$ pairs (where $t \in T$ and $s \in D(t)$) get a score from each activated score-based constraint (as explained before). That leaves us with up to 5 scores per pair that we would like to use to guide our solution exploration. To do so, we combine these into marginals, following in a way the example set by \cite{CEM} where the authors try to minimize a sum of energies. We compute these marginals $M(t,s)$ as the product of scores normalized over $D(t)$ :

    \begin{equation}
            M(t,s) = 
            \frac{\displaystyle
                \prod_{c} S_c(t,s)
                }
                {\displaystyle
                    \sum_{k\in D(t)} \prod_{c} S_c(t,k)
                }
    \end{equation}

Usually when Belief Propagation (\emph{BP}) is used in CP, the marginals built represent the density of solutions resulting from the branching of a constraint for the considered variable-value pair. It is used here to convey our marginals.

\subsubsection{Exploration method}
A branching heuristic called \emph{MaxMarginal} has been developed in \emph{MiniCPBP} \cite{miniCPBP}. It consists in guiding the construction of the search tree exploring the pairs tracklet-successor by descending order of marginal.
Regarding the exploration strategy, to prioritize staying close to the model by promoting high marginals association first, we use \emph{Depth First Search} ($DFS$) that consists in exploring the search tree by taking deviations as low as possible in the search tree if a valid is not found at first.

\subsection{Interpolation model}
\begin{figure}[!t]
        \centering
        \includegraphics[width=.4\textwidth]{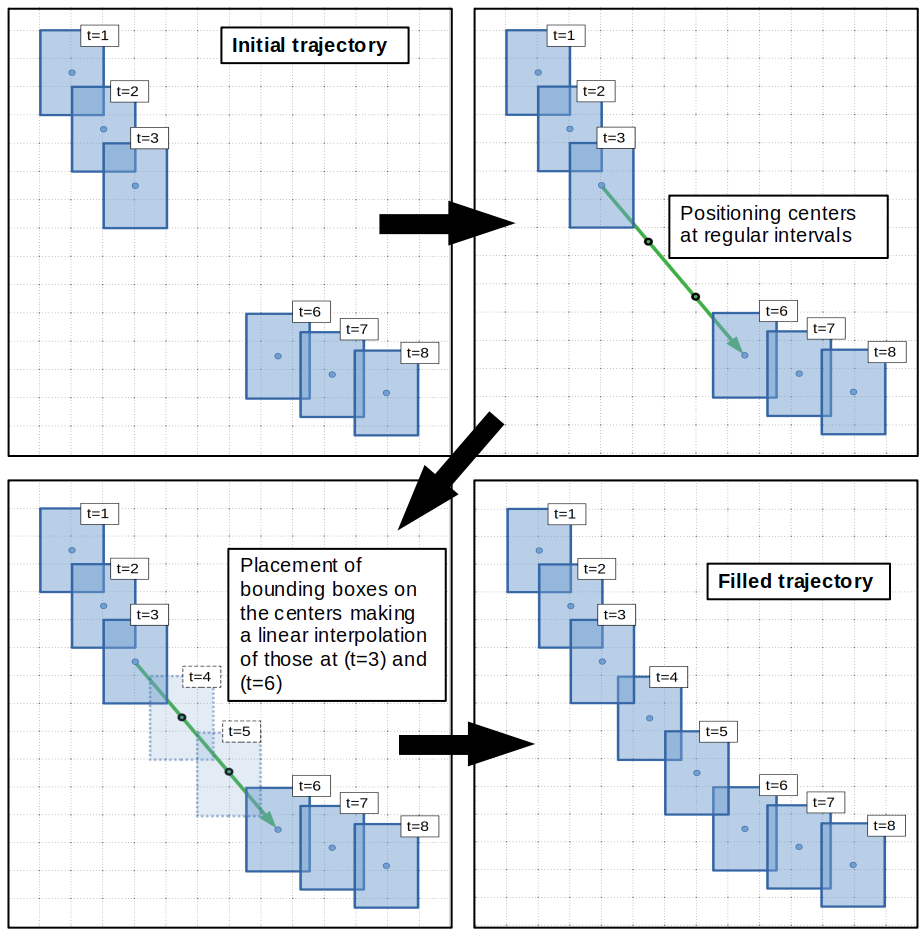}
        \caption{Workings of the interpolation model for $maxGapSize = 4$ : two detections are being placed to fill a hole, by placing centers regularly between the edges of the hole and then interpolating their dimensions linearly.}
        \label{fig:interpolation}
        
\end{figure}

Once the tracklets are associated with each other, it is likely that the resulting trajectories will have gaps, i.e. sequences of frames in which the object disappears before reappearing. This is why we decided to integrate a simple interpolation model in our method, which works as follows:

 As shown in Figure~\ref{fig:interpolation}, we identify the holes in each trajectory and if they are smaller than a threshold, $maxGapSize$, we fill them by making a linear interpolation from the detection preceding the hole to the one following it. Simply put, we consider that the object has moved (and changed shape or size) at a constant speed from the detection that precedes the hole to the one that follows it, and we add all the missing detections to the trajectory.

\section{Experiments and results}

\subsection{Evaluation Dataset}
We have chosen to evaluate our model on the training set of the \emph{MOT17} \cite{MOT16} challenge based on the detections of the best proposed detector, \emph{FRCNN} \cite{FRCNN}. This dataset represents a reference in the field and presents many difficulties that are particularly interesting to confront. Whether it is the often high occlusion rates, turbulence, moving cameras, strong variations in light exposure, sometimes subjective POV and other times very elevated POV, or the numerous reflections present in these videos, we are dealing with extremely varied situations that should allow us to evaluate our model in the majority of situations that can happen in urban settings.

\subsection{Evaluation metrics}
We evaluate our method using three of the main \emph{MOT} metrics:
\emph{MOTA} (and \emph{MOTP}) \cite{CLEAR-MOT} that mainly measure the quality of detections, and are used very broadly in the literature,
\emph{IDF1} \cite{IDF1} that refers more to the quality of the association between detections, also widely used in the literature,
and \emph{HOTA} \cite{HOTA} a more recent metric that accounts for both the performance in terms of association of detections and the quality of these detections.

\subsection{Parameters}

\begin{table}[!t]
\centering
\caption{Final model configurations used in ablation testing}
\resizebox{.5\textwidth}{!}{%
\begin{tabular}{|ll|l|}
\hline
\multicolumn{2}{|c|}{\textbf{Module}}                                                                   & \multicolumn{1}{c|}{\textbf{Best configuration}} \\ \hline
\multicolumn{1}{|l|}{\multirow{10}{*}{\textbf{Association}}} & \multirow{2}{*}{TimeDistance}            & $td_{50}$ = 1                                                    \\ \cline{3-3} 
\multicolumn{1}{|l|}{} &                                 & $td_{end}$ = 3      \\ \cline{2-3} 
\multicolumn{1}{|l|}{}                                       & \multirow{2}{*}{PredictedCenterDistance} & $pcd_{50}$ = 0.02                                                \\ \cline{3-3} 
\multicolumn{1}{|l|}{} &                                 & $pcd_{end}$ = 2     \\ \cline{2-3} 
\multicolumn{1}{|l|}{} & \multirow{2}{*}{PredictedIOU}   & $piou_{50}$ = 0.75  \\ \cline{3-3} 
\multicolumn{1}{|l|}{} &                                 & $piou_{end}$ = 2    \\ \cline{2-3} 
\multicolumn{1}{|l|}{} & AngleDifference        & \textit{disabled} \\ \cline{2-3} 
\multicolumn{1}{|l|}{} & SpeedNormDifference    & \textit{disabled} \\ \hline
\multicolumn{2}{|l|}{\textbf{TrackletCutter}}                                & $T_{TC}$ = 0.5                                                   \\ \hline
\multicolumn{2}{|l|}{\textbf{Interpolation}}             & $maxGapSize$ = 42   \\ \hline
\end{tabular}%
}
\label{table:config}
\end{table}
We tested multiple combinations of parameters for each of our modules (tried to activate or not each of the constraints) to find interactions and select the best configuration. Table \ref{table:config} represents the best configuration we found by applying our model to the \emph{MOT17} training data. We found during these calibration sessions that the vast majority of high scoring configurations were those that did not give the ability to filter successors (i.e. reduce their score to 0) to score-based constraints. We therefore disabled this ability of constraints that only guide the search in the configuration presented below.
 
Concerning the exploration of the solutions, it turned out that by pushing this one even up to the $10,000^{th}$ valid solution explored for multiple configurations of parameters and constraints, we did not obtain better results than by stopping at the first one found by branching on the maximal marginals (except for the very bad models, which obtained in all cases worst solutions than the initial tracker), so we decided to stop at the first valid solution in our exploration.
The code for our method can be found at \href{https://github.com/reminahon/tracklet\_associator}{github.com/reminahon/tracklet\_associator}.

\subsection{Results and Discussion}

\begin{table}[!t]
\caption{Results of the different model components in ablation on three different trackers applied to the \emph{MOT17} training sequences. TC: \emph{TrackletCutter}, CP: \emph{CP Associator}, Int: \emph{Interpolation model}. For all metrics, higher is better.}
\centering
\resizebox{.5\textwidth}{!}{%
\begin{tabular}{|l|l|l|l|}
\hline
\textbf{Method}            & \textbf{HOTA}    & \textbf{MOTA}    & \textbf{IDF1}    \\ \hline
IOU-Tracker                 & 43.04\%          & 49.74\%          & 50.27\%          \\
IOU-Tracker + CP            & 45.34\%          & 49.91\%          & 53.78\%          \\
IOU-Tracker + TC + CP       & 45.47\%          & 49.90\%          & 54.11\%          \\
IOU-Tracker + CP + Int      & 46.04\%          & \textbf{50.62\%} & 54.38\%          \\
IOU-Tracker + TC + CP + Int & \textbf{46.18\%} & 50.49\%          & \textbf{54.74\%} \\ \hline
SORT                        & 42.80\%          & 48.54\%          & 50.63\%          \\
SORT + CP                   & 45.40\%          & 48.71\%          & 54.34\%          \\
SORT + TC + CP              & 45.14\%          & 48.70\%          & 53.96\%          \\
SORT + CP + Int             & \textbf{46.35\%} & \textbf{49.46\%} & \textbf{54.92\%} \\
SORT + TC + CP + Int        & 46.06\%          & 49.39\%          & 54.58\%          \\ \hline
Tracktor                    & 55.18\%          & 61.81\%          & 65.06\%          \\
Tracktor + CP               & 56.39\%          & 61.87\%          & 67.40\%          \\
Tracktor + TC + CP          & 55.89\%          & 61.85\%          & 66.77\%          \\
Tracktor + CP + Int         & \textbf{56.98\%} & \textbf{62.99\%} & \textbf{67.88\%} \\
Tracktor + TC + CP + Int    & 56.49\%          & 62.96\%          & 67.29\%     \\ \hline
\end{tabular}%
}
\label{table:Results}
\end{table}

Results are given in Table \ref{table:Results}. 
It can be noted that whatever the tracker we apply it to, our model allows to obtain improvements of several percents on the three scores that interest us: \emph{HOTA}, \emph{MOTA} and \emph{IDF1}. Concerning the \emph{HOTA}, the main metric of our evaluation, we obtain an improvement of $3.14\%$ for \emph{IOU-Tracker}, $3.55\%$ for \emph{SORT} and $1.8\%$ for \emph{Tracktor} which already had a rather high score ($13\%$ more than the two others originally) which shows that our module is likely effective on any type of tracker independently of their initial level of performance or their tracking paradigm. 

Our goal was mainly to improve data association with our \emph{CP Associator}, but we also addressed the detection phase with the \emph{interpolation} module. \emph{HOTA} and \emph{IDF1} are the two metrics that are the most sensitive to the quality of the associations, as opposed to \emph{MOTA} that is not very sensitive to the data association quality, but more sensitive to the detections quality. We observe that the \emph{CP} association model is the module that allows the most improvements in terms of \emph{HOTA} and \emph{IDF1} ($70\%$ of the improvements on average) to the results of the three trackers.  The rest of the improvements are mainly brought by the interpolation model which allows to improve the MOTA by more than one point for \emph{Tracktor}, in particular, by adding missing detections. 

However, the \emph{IOU-Tracker} is the only tracker for which the \emph{TrackletCutter} really allows an improvement of the results. This may be due to the fact that this tracker has more errors due to occlusions detected by our method. Still, it seems that our model performs adequately without the \emph{TrackletCutter}. It turns out that this is the part of the model that requires the most computation time, for little to no improvement. 
We would therefore advise not to use it for any other tracker than \emph{IOU-Tracker}. Moreover, on the whole MOT17 training set, applying our model to get the improved trajectories takes between 30 to 60 seconds without the \emph{TrackletCutter} and up to 5 minutes with it. Even without the use of the \emph{TrackletCutter}, our model retains interest insofar as trackers tend to suffer from fragmentation which we correct by our association. One could even postulate that the more a tracker suffers from fragmentation, the better our post-processing can help its tracking performance.

\section{Conclusion}

We presented a method that can be used as a post-processing step for any state-of-the-art multi-object tracker to improve its association performance as we have been able to show by testing it on the trackers \emph{IOU-Tracker}, \emph{SORT} and \emph{Tracktor} on the \emph{MOT17} dataset. This demonstrates its competitiveness in the field of pedestrian tracking. In addition we propose here the first association model based on Constraint Programming with Belief Propagation. Furthermore, a strength of our method for future improvements relies on our modularity: each module we propose (whether it is the \emph{TrackletCutter}, the association model or the interpolation one) can be substituted with another one that would accomplish the same function. New constraints based on other characteristics (such as appearance for example) can also be added without any major changes in the architecture of the model.






\bibliographystyle{IEEEtran}
\bibliography{IEEEabrv, bare_conf.bib}

\begin{thebibliography}{10}
\providecommand{\url}[1]{#1}
\csname url@samestyle\endcsname
\providecommand{\newblock}{\relax}
\providecommand{\bibinfo}[2]{#2}
\providecommand{\BIBentrySTDinterwordspacing}{\spaceskip=0pt\relax}
\providecommand{\BIBentryALTinterwordstretchfactor}{4}
\providecommand{\BIBentryALTinterwordspacing}{\spaceskip=\fontdimen2\font plus
\BIBentryALTinterwordstretchfactor\fontdimen3\font minus
  \fontdimen4\font\relax}
\providecommand{\BIBforeignlanguage}[2]{{%
\expandafter\ifx\csname l@#1\endcsname\relax
\typeout{** WARNING: IEEEtran.bst: No hyphenation pattern has been}%
\typeout{** loaded for the language `#1'. Using the pattern for}%
\typeout{** the default language instead.}%
\else
\language=\csname l@#1\endcsname
\fi
#2}}
\providecommand{\BIBdecl}{\relax}
\BIBdecl

\bibitem{RCNN}
R.~Girshick, J.~Donahue, T.~Darrell, and J.~Malik, ``Rich feature hierarchies
  for accurate object detection and semantic segmentation,'' in \emph{2014 IEEE
  Conference on Computer Vision and Pattern Recognition}, 2014, pp. 580--587.

\bibitem{FRCNN}
\BIBentryALTinterwordspacing
S.~Ren, K.~He, R.~B. Girshick, and J.~Sun, ``Faster {R-CNN:} towards real-time
  object detection with region proposal networks,'' \emph{CoRR}, vol.
  abs/1506.01497, 2015. [Online]. Available:
  \url{http://arxiv.org/abs/1506.01497}
\BIBentrySTDinterwordspacing

\bibitem{IOU2017}
E.~Bochinski, V.~Eiselein, and T.~Sikora, ``High-speed tracking-by-detection
  without using image information,'' in \emph{2017 14th IEEE International
  Conference on Advanced Video and Signal Based Surveillance (AVSS)}, 2017, pp.
  1--6.

\bibitem{Kalman}
R.~E. Kalman, ``{A New Approach to Linear Filtering and Prediction Problems},''
  \emph{Journal of Basic Engineering}, vol.~82, no.~1, pp. 35--45, 03 1960.

\bibitem{SORT}
A.~Bewley, Z.~Ge, L.~Ott, F.~Ramos, and B.~Upcroft, ``Simple {Online} and
  {Realtime} {Tracking},'' \emph{2016 IEEE International Conference on Image
  Processing (ICIP)}, pp. 3464--3468, Sep. 2016, arXiv: 1602.00763.

\bibitem{CMC}
G.~D. Evangelidis and E.~Z. Psarakis, ``Parametric image alignment using
  enhanced correlation coefficient maximization,'' \emph{IEEE Transactions on
  Pattern Analysis and Machine Intelligence}, vol.~30, no.~10, pp. 1858--1865,
  2008.

\bibitem{HOC}
C.~Novak and S.~Shafer, ``Anatomy of a color histogram,'' in \emph{Proceedings
  1992 IEEE Computer Society Conference on Computer Vision and Pattern
  Recognition}, 1992, pp. 599--605.

\bibitem{HOG}
N.~Dalal and B.~Triggs, ``Histograms of oriented gradients for human
  detection,'' \emph{Computer Vision and Pattern Recognition, 2005. CVPR 2005.
  IEEE Computer Society Conference on}, vol.~1, pp. 886--893, 2005.

\bibitem{KCF}
J.~F. Henriques, R.~Caseiro, P.~Martins, and J.~Batista, ``High-speed tracking
  with kernelized correlation filters,'' \emph{IEEE Transactions on Pattern
  Analysis and Machine Intelligence}, vol.~37, no.~3, p. 583–596, Mar 2015.

\bibitem{DAN}
S.~Sun, N.~Akhtar, H.~Song, A.~Mian, and M.~Shah, ``Deep {Affinity} {Network}
  for {Multiple} {Object} {Tracking},'' \emph{arXiv:1810.11780 [cs]}, Jul.
  2019, arXiv: 1810.11780.

\bibitem{MOTer_TransCtr}
Y.~Xu, Y.~Ban, G.~Delorme, C.~Gan, D.~Rus, and X.~Alameda-Pineda,
  ``{TransCenter}: {Transformers} with {Dense} {Queries} for
  {Multiple}-{Object} {Tracking},'' \emph{arXiv:2103.15145 [cs]}, Mar. 2021,
  arXiv: 2103.15145.

\bibitem{trajeocc}
\BIBentryALTinterwordspacing
A.~Girbau, X.~Gir{\'{o}}{-}i{-}Nieto, I.~Rius, and F.~Marqu{\'{e}}s, ``Multiple
  object tracking with mixture density networks for trajectory estimation,''
  \emph{CoRR}, vol. abs/2106.10950, 2021. [Online]. Available:
  \url{https://arxiv.org/abs/2106.10950}
\BIBentrySTDinterwordspacing

\bibitem{Tracktor}
P.~Bergmann, T.~Meinhardt, and L.~Leal-Taixe, ``Tracking without bells and
  whistles,'' \emph{2019 IEEE/CVF International Conference on Computer Vision
  (ICCV)}, pp. 941--951, Oct. 2019, arXiv: 1903.05625.

\bibitem{CenterTrack}
X.~Zhou, V.~Koltun, and P.~Krähenbühl, ``Tracking {Objects} as {Points},''
  \emph{arXiv:2004.01177 [cs]}, Aug. 2020, arXiv: 2004.01177.

\bibitem{GOG}
H.~Pirsiavash, D.~Ramanan, and C.~C. Fowlkes,
  ``\BIBforeignlanguage{en}{Globally-optimal greedy algorithms for tracking a
  variable number of objects},'' in \emph{\BIBforeignlanguage{en}{{CVPR}
  2011}}.\hskip 1em plus 0.5em minus 0.4em\relax Colorado Springs, CO, USA:
  IEEE, Jun. 2011, pp. 1201--1208.

\bibitem{TMOH}
D.~Stadler and J.~Beyerer, ``Improving multiple pedestrian tracking by track
  management and occlusion handling,'' in \emph{2021 IEEE/CVF Conference on
  Computer Vision and Pattern Recognition (CVPR)}, 2021, pp. 10\,953--10\,962.

\bibitem{H2T}
L.~Wen, W.~Li, J.~Yan, Z.~Lei, D.~Yi, and S.~Z. Li, ``Multiple {Target}
  {Tracking} {Based} on {Undirected} {Hierarchical} {Relation} {Hypergraph},''
  in \emph{2014 {IEEE} {Conference} on {Computer} {Vision} and {Pattern}
  {Recognition}}, Jun. 2014, pp. 1282--1289, iSSN: 1063-6919.

\bibitem{pineault_article}
\BIBentryALTinterwordspacing
A.~Pineault, G.~Bilodeau, and G.~Pesant, ``Tracking road users using constraint
  programming,'' \emph{CoRR}, vol. abs/2003.04468, 2020. [Online]. Available:
  \url{https://arxiv.org/abs/2003.04468}
\BIBentrySTDinterwordspacing

\bibitem{pesant_cp}
G.~Pesant, ``\BIBforeignlanguage{en}{A constraint programming primer},''
  \emph{\BIBforeignlanguage{en}{EURO Journal on Computational Optimization}},
  vol.~2, no.~3, pp. 89--97, Aug. 2014.

\bibitem{miniCPBP}
------, ``\BIBforeignlanguage{en}{From {Support} {Propagation} to {Belief}
  {Propagation} in {Constraint} {Programming} ({Extended} {Abstract})},'' in
  \emph{\BIBforeignlanguage{en}{Proceedings of the {Twenty}-{Ninth}
  {International} {Joint} {Conference} on {Artificial} {Intelligence}}}, Jul.
  2020, pp. 5100--5104.

\bibitem{CEM}
A.~Milan, S.~Roth, and K.~Schindler, ``\BIBforeignlanguage{en}{Continuous
  {Energy} {Minimization} for {Multitarget} {Tracking}},''
  \emph{\BIBforeignlanguage{en}{IEEE Transactions on Pattern Analysis and
  Machine Intelligence}}, vol.~36, no.~1, pp. 58--72, Jan. 2014.

\bibitem{MOT16}
\BIBentryALTinterwordspacing
A.~Milan, L.~Leal{-}Taix{\'{e}}, I.~D. Reid, S.~Roth, and K.~Schindler,
  ``{MOT16:} {A} benchmark for multi-object tracking,'' 2016. [Online].
  Available: \url{http://arxiv.org/abs/1603.00831}
\BIBentrySTDinterwordspacing

\bibitem{CLEAR-MOT}
K.~Bernardin and R.~Stiefelhagen, ``Evaluating {Multiple} {Object} {Tracking}
  {Performance}: {The} {CLEAR} {MOT} {Metrics},'' \emph{EURASIP Journal on
  Image and Video Processing}, vol. 2008, no.~1, p. 246309, 2008.

\bibitem{IDF1}
E.~Ristani, F.~Solera, R.~S. Zou, R.~Cucchiara, and C.~Tomasi, ``Performance
  {Measures} and a {Data} {Set} for {Multi}-{Target}, {Multi}-{Camera}
  {Tracking},'' \emph{arXiv:1609.01775 [cs]}, Sep. 2016, arXiv: 1609.01775.

\bibitem{HOTA}
J.~Luiten, A.~Osep, P.~Dendorfer, P.~Torr, A.~Geiger, L.~Leal-Taixé, and
  B.~Leibe, ``{HOTA}: {A} {Higher} {Order} {Metric} for {Evaluating}
  {Multi}-object {Tracking},'' \emph{International Journal of Computer Vision},
  vol. 129, no.~2, pp. 548--578, 2021.

\end{thebibliography}
%

\end{document}